\documentclass[letterpaper]{article} 
\usepackage{aaai24}  
\usepackage{times}  
\usepackage{helvet}  
\usepackage{courier}  
\usepackage[hyphens]{url}  
\usepackage{graphicx} 
\usepackage{natbib}  
\usepackage{caption} 
\usepackage{algorithm}
\usepackage{algorithmic}
\usepackage{amsmath}
\usepackage{amssymb}

\usepackage{amsthm}

\usepackage{subcaption}
\usepackage{booktabs}
\usepackage{multirow}

\usepackage{newfloat}
\usepackage{listings}
\DeclareCaptionStyle{ruled}{labelfont=normalfont,labelsep=colon,strut=off} 
\floatstyle{ruled}
\newfloat{listing}{tb}{lst}{}
\floatname{listing}{Listing}
\title{Domain-Aware Fine-Tuning: Enhancing Neural Network Adaptability}

\author {
    Seokhyeon Ha\textsuperscript{\rm1},
    Sunbeom Jeong\textsuperscript{\rm1},
    Jungwoo Lee\textsuperscript{\rm1,2}
}
\affiliations {
    \textsuperscript{\rm 1} Seoul National University \\
    \textsuperscript{\rm 2} HodooAI Lab  \\
    \{aoxm1231, sb3991, junglee\}@snu.ac.kr    
}

\usepackage{bibentry}

\begin{document}

\maketitle

\begin{abstract}
Fine-tuning pre-trained neural network models has become a widely adopted approach across various domains. However, it can lead to the distortion of pre-trained feature extractors that already possess strong generalization capabilities. Mitigating feature distortion during adaptation to new target domains is crucial. Recent studies have shown promising results in handling feature distortion by aligning the head layer on in-distribution datasets before performing fine-tuning. Nonetheless, a significant limitation arises from the treatment of batch normalization layers during fine-tuning, leading to suboptimal performance. In this paper, we propose Domain-Aware Fine-Tuning (DAFT), a novel approach that incorporates batch normalization conversion and the integration of linear probing and fine-tuning. Our batch normalization conversion method effectively mitigates feature distortion by reducing modifications to the neural network during fine-tuning. Additionally, we introduce the integration of linear probing and fine-tuning to optimize the head layer with gradual adaptation of the feature extractor. By leveraging batch normalization layers and integrating linear probing and fine-tuning, our DAFT significantly mitigates feature distortion and achieves improved model performance on both in-distribution and out-of-distribution datasets. Extensive experiments demonstrate that our method outperforms other baseline methods, demonstrating its effectiveness in not only improving performance but also mitigating feature distortion.
\end{abstract}

\section{Introduction}

\begin{figure*}
     \centering
     \begin{subfigure}[b]{0.45\textwidth}         
         \includegraphics[width=\textwidth]{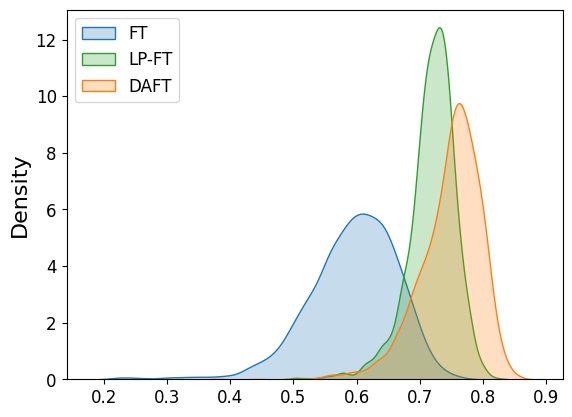}
         \caption{Cosine Similarity}
        \label{fig:fmow_feature_CosSimilarity}
     \end{subfigure}
     \hfill
     \begin{subfigure}[b]{0.45\textwidth}
         
         \includegraphics[width=\textwidth]{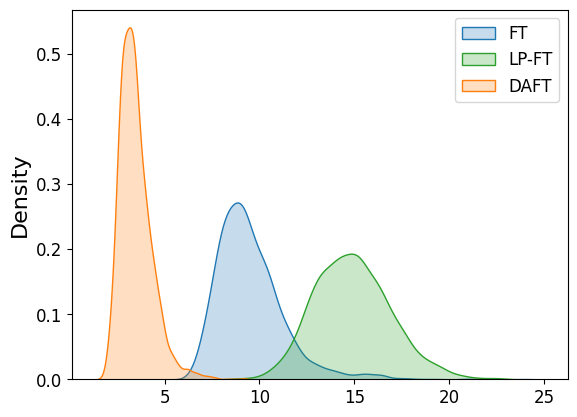}
         \caption{L2 Distance}
        \label{fig:fmow_feature_L2Distance}
     \end{subfigure}     
     \caption{
    Distribution of similarity measures for three different fine-tuning methods. We fine-tune the pre-trained ResNet-50 model from MoCo-v2 \cite{chen2020improved} on fMoW \cite{christie2018functional} dataset. First, we extract features from the test data before applying each fine-tuning method. Then, after completing the fine-tuning process, we compute the similarity measures between the pre-fine-tuning and post-fine-tuning features for each method. The similarity measures, including cosine similarity and L2 distance, are computed on the fMoW test dataset. Notably, our DAFT exhibits the least distortion in the pre-trained features across both similarity measures.
     }
     \label{fig:fmow_feature_histogram_measures}
\end{figure*}

Transferable neural network models have seen significant advancements in various domains, including computer vision \cite{he2020momentum, chen2020simple, radford2021learning}, natural language processing \cite{kenton2019bert, yang2019xlnet}, and speech recognition \cite{baevski2020wav2vec, gulati2020conformer}. These models, trained on large-scale datasets with substantial computational resources, demonstrate exceptional performance and generalization capabilities, making them widely used in transfer learning. By employing these pre-trained models, transfer learning can substantially reduce the time and resources required for training and data collection, while also enhancing performance compared to training from scratch.

When transferring knowledge from a pre-trained model, the feature extractor is first initialized with the pre-trained model. Subsequently, there are two primary optimization methods commonly employed: linear probing (LP) and fine-tuning (FT). In LP, only the head layer is optimized, while the feature extractor is fixed during training. On the other hand, FT involves optimizing both the feature extractor and the linear head layer simultaneously. While FT generally outperforms LP on in-distribution (ID) datasets, recent findings show that FT performs worse on out-of-distribution (OOD) datasets compared to LP. It is due to feature distortion in pre-trained features that possess good generalization capabilities. To mitigate feature distortion, LP-FT was proposed as a two-stage training approach \cite{kumar2022fine}. In the first stage, LP aligns the head layer with the ID subspace, and in the second stage, FT is performed with a small learning rate to mitigate feature distortion. This allows LP-FT to achieve improved performance on both ID and OOD datasets.

However, LP-FT still has several limitations. One of the key issues is the treatment of batch normalization layers during the FT stage in the LP-FT framework. During the FT stage, all batch normalization layers are switched to test mode, where fixed statistics are employed instead of batch statistics. While this prevents the alteration of batch normalization statistics during training, potentially mitigating feature distortion, it also harms beneficial effects such as improved gradient flow and reduced internal covariate shift \cite{ioffe2015batch}. Consequently, the overall optimization process is degraded, leading to suboptimal model performance. Furthermore, the initial LP stage in LP-FT optimizes the head layer with a feature extractor that has not been adapted to the target domain, which might not be ideal for domain adaptation. Moreover, LP-FT utilizes batch normalization with fixed test statistics instead of batch statistics, which can lead to gradient exploding if not accompanied by a significantly small learning rate. Therefore, LP-FT employs a smaller learning rate during the FT stage compared to standard FT. It limits the optimization of the head layer during the FT stage, especially when there is a significant dissimilarity in data distribution between the source and target domains. These limitations call for a more effective approach that can adequately address feature distortion while effectively performing batch normalization.

To address these challenges, we present a novel method named Domain-Aware Fine-Tuning (DAFT). Our method is built upon two core techniques: batch normalization conversion and the integration of LP and FT into a single stage. The batch normalization conversion method modifies batch normalization layers to better suit the target domain prior to fine-tuning. It makes batch normalization more effective, reducing feature distortion by adjusting the statistics and parameters to the target domain. In addition, our DAFT integrates LP and FT into a single stage, where the linear head is optimized with gradually adapting features from the target domain. This strategy mitigates the issue of the head layer being optimized solely using a feature extractor that has not yet adapted to the target domain, resulting in improved optimization.

Through these techniques, our DAFT effectively reduces feature distortion and achieves enhanced performance compared to other baseline methods. The effectiveness of our method is demonstrated through similarity measures computed between features before and after adaptation. As shown in Figure \ref{fig:fmow_feature_histogram_measures}, our DAFT consistently exhibits higher feature similarity compared to existing methods such as FT and LP-FT. Furthermore, we conduct extensive experiments to compare our proposed method against other baseline methods. These results demonstrate the superior performance of the proposed DAFT with higher accuracy and less feature distortion.

In summary, our main contributions are as follows.
\begin{itemize}
\item We propose a novel batch normalization conversion technique, which improves batch normalization during fine-tuning with reducing modifications to the neural network.
\item The integration of LP and FT is proposed, allowing the head layer to be optimized with the gradual adaptation of the feature extractor.
\item Through extensive experiments, we demonstrate the superior performance of our proposed method compared to other baseline approaches.
\end{itemize}

\section{Related Works}
\subsubsection{Head Initialization for Fine-Tuning} LP-FT, a two-stage transfer learning method, initializes the head layer with parameters obtained from LP prior to executing FT \cite{kumar2022fine}. This highlights that random initialization of the head layer has the potential to cause feature distortion in FT, leading to degraded generalization performance. Nonetheless, the performance of LP-FT remains limited even with aligned head initialization, calling for further research on better head initialization methods. One promising technique is to stop the LP stage early before it reaches convergence, resulting in a non-converged head that improves performance during the subsequent FT stage \cite{ren2023prepare}. Additionally, applying hardness-promoting augmentation during the LP stage can help mitigate feature distortion and also simplicity bias, leading to enhanced generalization performance \cite{trivedi2023closer}. Although these head initialization techniques have shown effectiveness in boosting the performance of the FT stage, they require a separate training stage to initialize the head layer before the FT stage. In contrast, our proposed DAFT integrates head adaptation seamlessly without the need for a distinct initialization stage.

\subsubsection{Other Modifications for Fine-Tuning}
There are several other techniques to improve the performance of FT. FLYP utilizes contrastive learning with additional text data \cite{goyal2023finetune}. Side-tuning involves fixing the feature extractor to preserve pre-trained features and combining it with a small side network, then training only the side network and the head layer \cite{zhang2020side}. Other methods combine FT with the regularization of pre-trained parameters \cite{xuhong2018explicit, li2019delta, gouk2021distance, li2021improved}, or the use of different learning rates for each layer \cite{howard2018universal, yang2019xlnet, clark2020electra}. Our method falls under the category of employing different learning rates, as we use a larger learning rate for the head layer. However, unlike general methods that train the head layer with a learning rate approximately 10 times larger than that of the feature extractor, our method uses separate learning rates for the head layer and the feature extractor. This allows the head layer to align better with the ID dataset during the early stage of training and to converge with the adapting feature extractor.

\subsubsection{Batch Normalization in Transfer Learning} Batch normalization is a well-known technique for covariance shift reduction and better stable training \cite{ioffe2015batch}. However, in transfer learning scenarios, where the data distribution between the source domain and target domain may significantly differ, the movement of statistics within batch normalization layers can result in severe feature distortion. As a solution, many methods choose to freeze the batch normalization statistics during fine-tuning to alleviate feature distortion \cite{kumar2022fine, ren2023prepare, trivedi2023closer}. On the other hand, some methods leverage batch normalization with domain-specific statistics to encourage the learning of more generalized features \cite{wang2019transferable, chang2019domain} or even use statistics from test batches to enhance robustness against domain shift \cite{mirza2022norm, lim2023ttn}. In contrast to existing approaches, our method aims to mitigate distortion of pre-trained features while still benefiting from batch normalization during training.

\section{Method}
In this section, we introduce our approach, Domain-Aware Fine-Tuning (DAFT), which incorporates converting batch normalization and integrating LP and FT.

\subsection{Converting Batch Normalization}
\subsubsection{Batch Normalization (BN)}

The Batch Normalization (BN) technique \cite{ioffe2015batch} was introduced based on the observation that network training tends to converge faster when its inputs are whitened \cite{lecun2002efficient, wiesler2011convergence}. BN operates in two modes: training mode and test mode. In the training mode, it first computes batch statistics, $\mu$ and $\sigma^{2}$, for the input features $z$. Then, scaling and shifting parameters, $\gamma$ and $\beta$, are applied to scale and shift the normalized values. The normalization process is represented as follows:

\begin{equation}
\label{eq:BN}
\hat{y} = \gamma\cdot\frac
{z - \mu }
{\sqrt{\sigma^{2}+ \epsilon}}+ \beta,
\end{equation}
where $\epsilon$ is a small constant for numerical stability. During training, BN computes the mean $\mu$ and variance $\sigma^{2}$ over a mini-batch of training data as
\begin{equation}
    \label{eq:batch_statistics}
    \mu = \frac{1}{m}\sum^{m}_{i=1}{z_{i}}, \quad \sigma^{2} =\frac{1}{m}\sum^{m}_{i=1}{(z_{i}-\mu)^{2}},
\end{equation}
where $m$ is the batch size. In the test mode, BN uses test statistics estimates, $\mathrm{M}$ and $\mathrm{\Sigma}$, instead of $\mu$ and $\sigma$. Test statistics are obtained by using the moving averages of the train batch statistics. These moving averages are continuously updated during training to ensure consistent normalization during the test mode without the need for additional statistics computation.

\begin{figure*}
    \centering
        \begin{subfigure}[b]{0.31\textwidth}         
         \includegraphics[width=\textwidth]{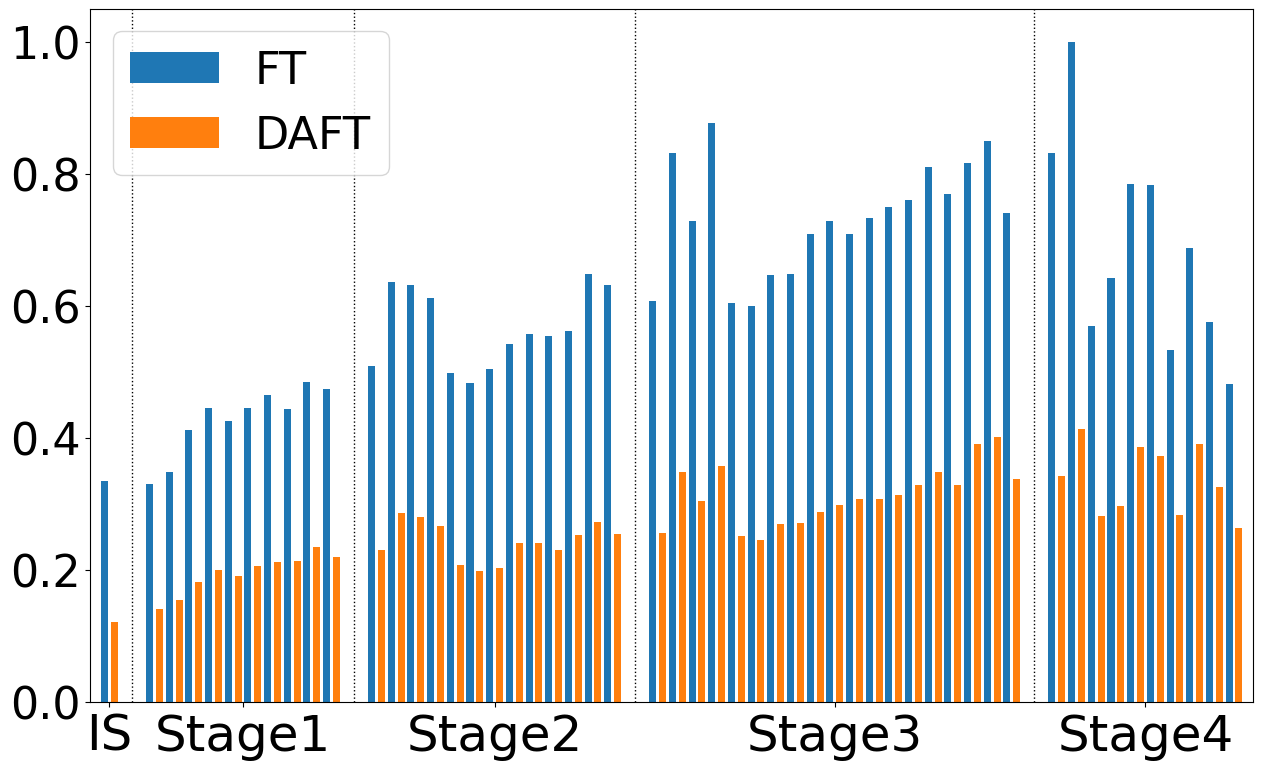}
         \caption{Weight of Conv Layers}
        \label{fig:conv_w}
        \end{subfigure}
    \quad
        \begin{subfigure}[b]{0.31\textwidth}         
         \includegraphics[width=\textwidth]{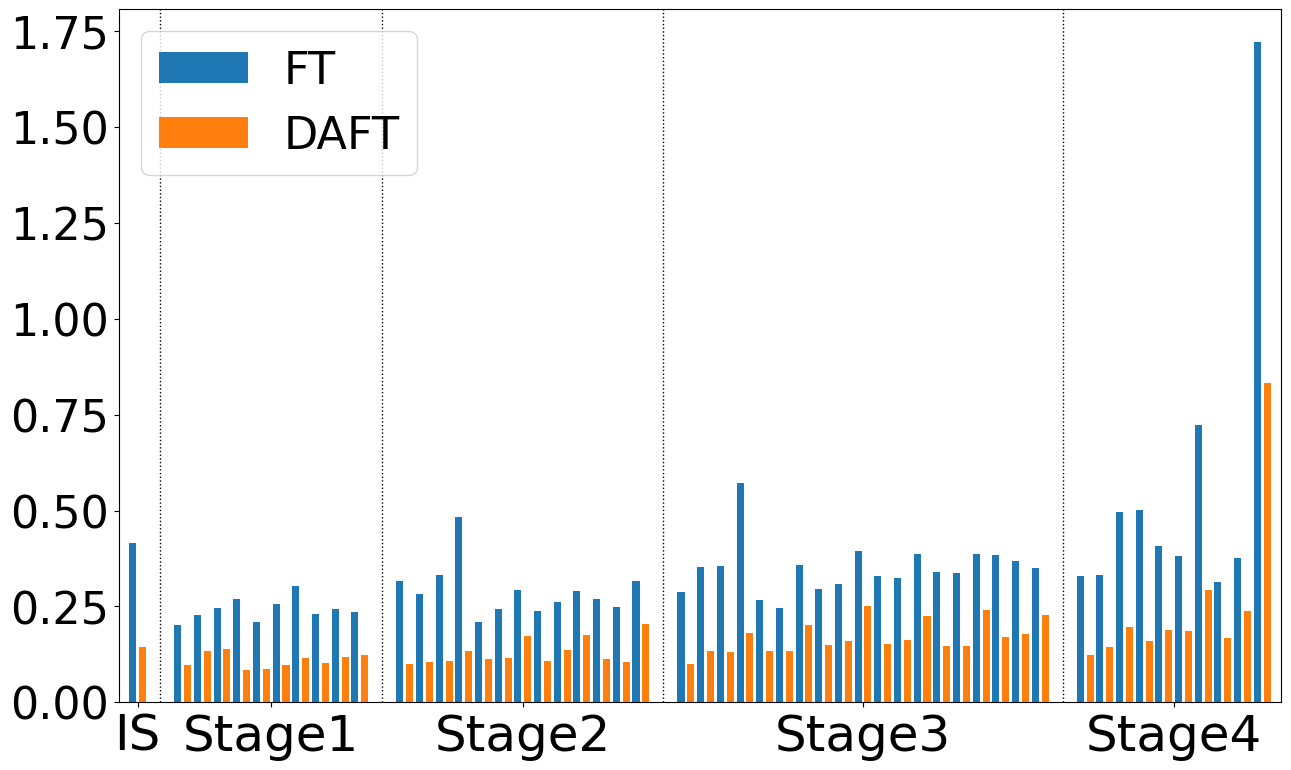}
         \caption{Weight of BN Layers}
        \label{fig:bn_w}
        \end{subfigure}
    \quad
        \begin{subfigure}[b]{0.31\textwidth}         
         \includegraphics[width=\textwidth]{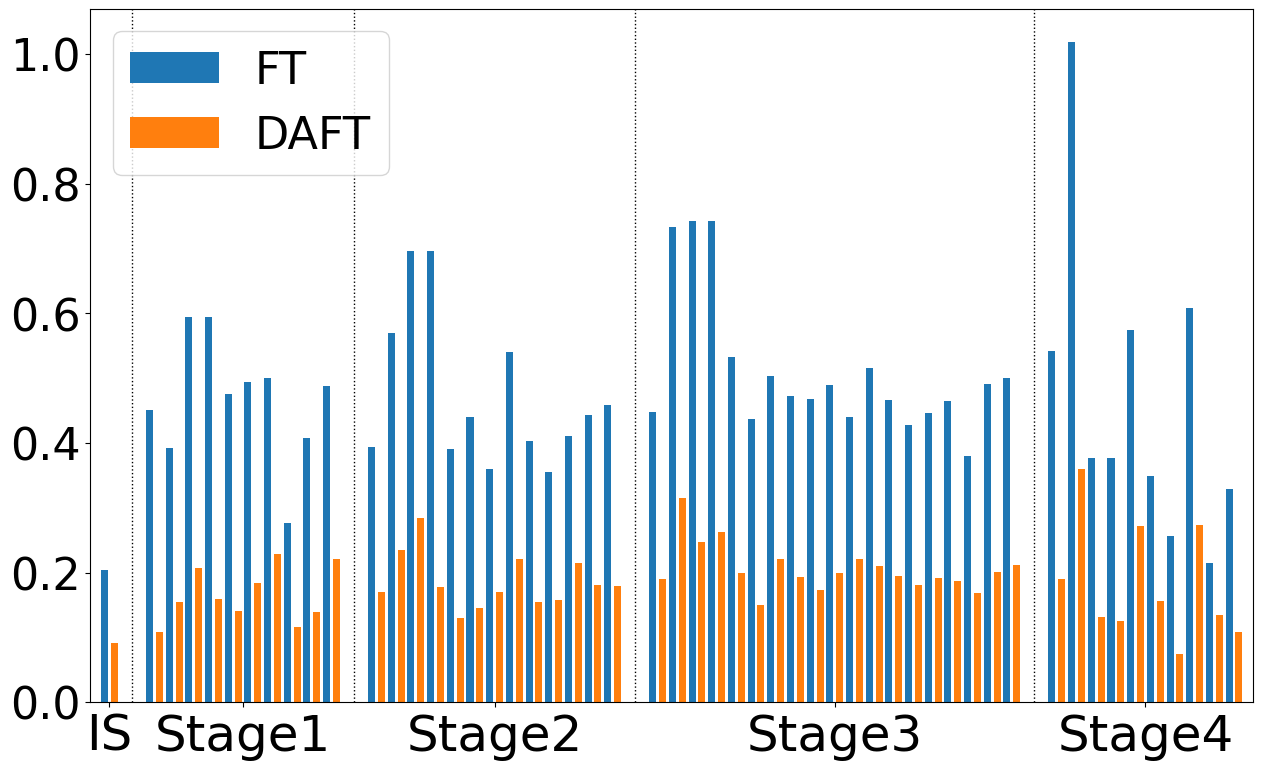}
         \caption{Bias of BN Layers}
        \label{fig:bn_b}
        \end{subfigure}
    \medskip
        \begin{subfigure}[b]{0.31\textwidth}         
         \includegraphics[width=\textwidth]{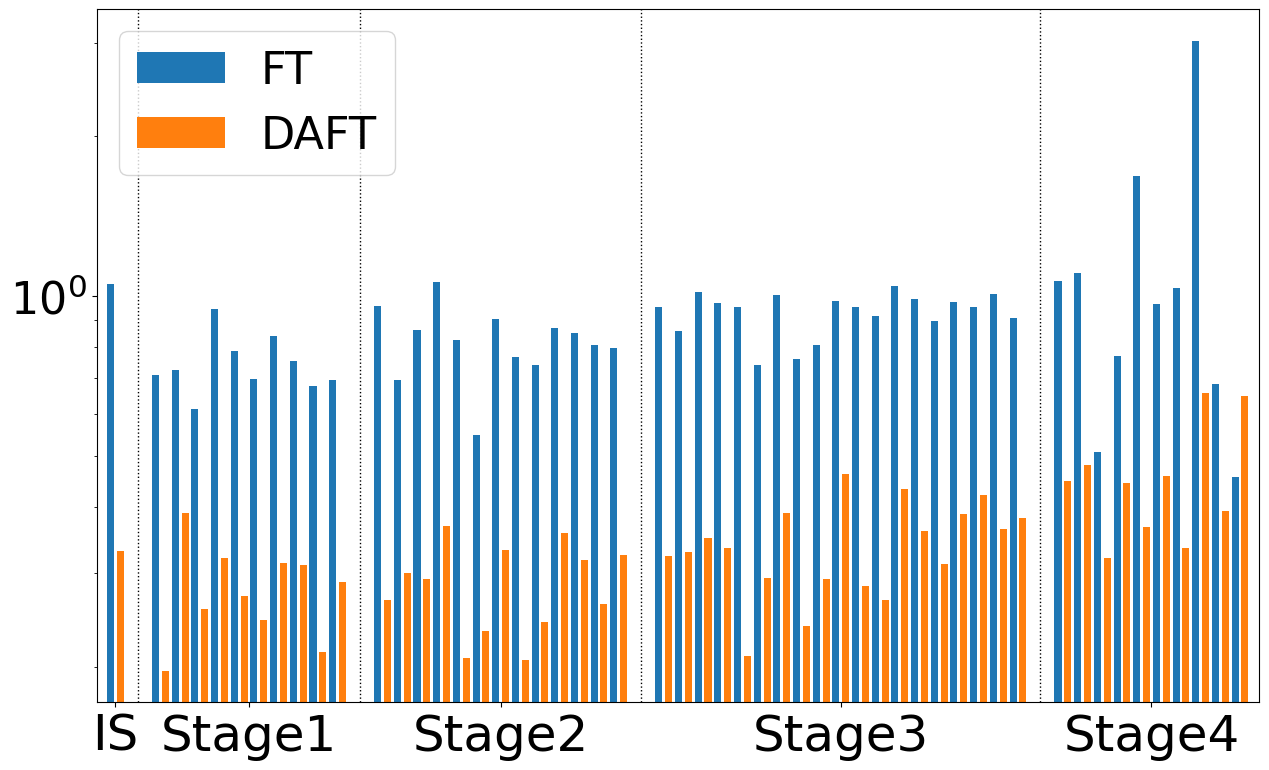}
         \caption{Mean of BN Layers}
        \label{fig:bn_m}
        \end{subfigure}
    \quad     
        \begin{subfigure}[b]{0.31\textwidth}         
         \includegraphics[width=\textwidth]{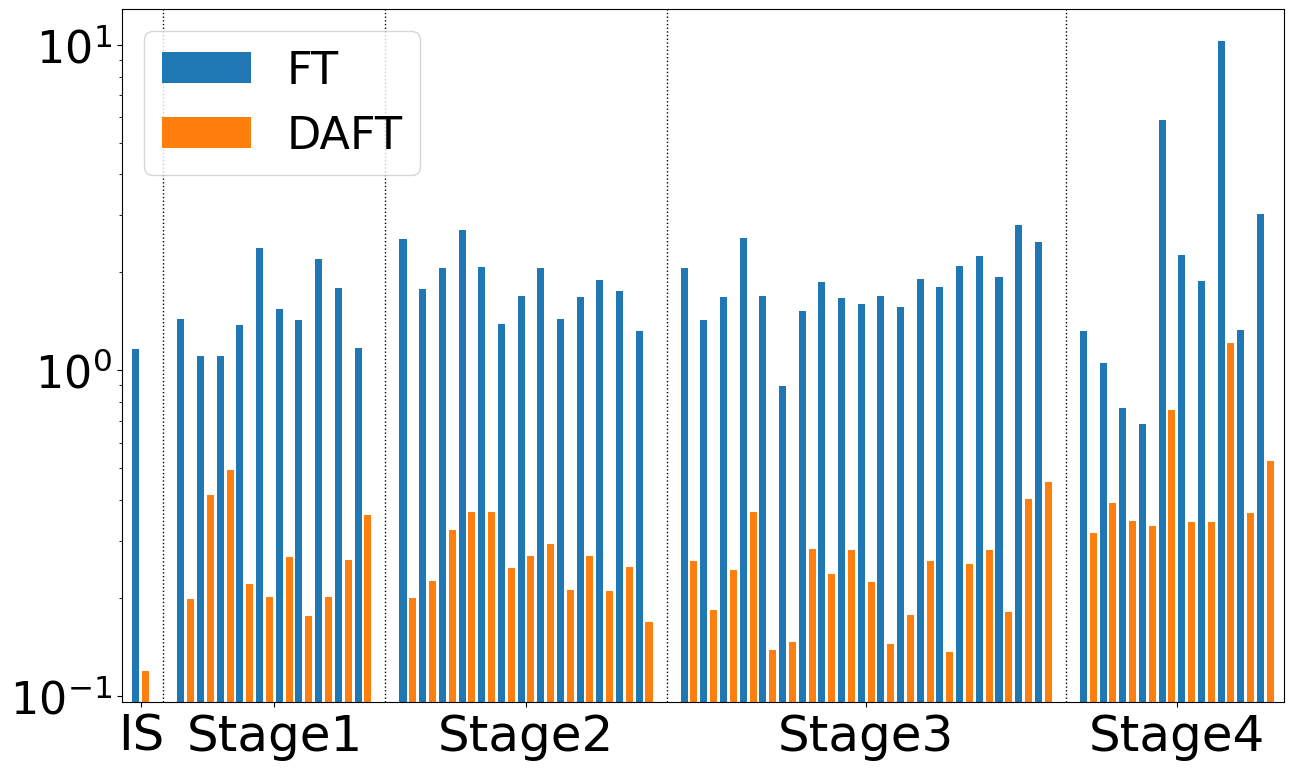}
         \caption{Variance of BN Layers}
        \label{fig:bn_v}
        \end{subfigure}    
    \caption{   
    Comparison of relative changes in learning parameters (\ref{fig:conv_w}, \ref{fig:bn_w}, \ref{fig:bn_b}) and BN statistics (\ref{fig:bn_m}, \ref{fig:bn_v}) between FT and our DAFT on the fMoW dataset. We utilize the pre-trained ResNet-50 model from MoCo-v2 as the feature extractor and conduct each fine-tuning method. The ResNet-50 architecture consists of an Input Stem and 4 subsequent stages, with each stage indicated on the x-axis from left to right. `IS' on the x-axis represents the Input Stem, and all layers within each stage are further indicated on the x-axis sequentially from left to right. The relative changes are computed between the initial values before each fine-tuning and the final values after each fine-tuning process. Note that the relative changes of BN statistics are represented in log scale.
    }
    \label{fig:relative_changes}
\end{figure*}

\subsubsection{BN Transfer Issue}
Due to the presence of \textit{dataset bias} or \textit{domain shift} \cite{quinonero2008dataset}, significant differences arise in the input distribution of batch normalization between the source and the target domains. Let $\mathrm{M}{s}$ and $\mathrm{\Sigma}{s}$ respectively represent the test mean and test variance of batch normalization, pre-trained on the source domain. When training a pre-trained model on a new target domain, the values of $\mathrm{M}_{s}$ and $\mathrm{\Sigma}_{s}$ undergo considerable updates, leading to a significant deviation from their original values. Moreover, the learning parameters also experience substantial adjustments during training to adapt and converge with the target domain statistics. These significant changes in batch normalization are one of the main factors that cause feature distortion during fine-tuning.

\subsubsection{Batch Normalization Conversion}
While many recent studies have addressed the BN transfer issue by introducing additional statistics and parameters for the new domain \cite{chang2019domain, lim2023ttn}, we handle this issue by simply converting the existing statistics and parameters without introducing additional ones. We introduce our BN conversion to effectively prevent the distortion of pre-trained feature extractors. Before starting the training process, we first compute batch statistics, $\mu_{t}$ and $\sigma_{t}$, for each mini-batch of the target training data. Next, we compute the new unbiased statistics, $\mathrm{M}_{t}$ and $\mathrm{\Sigma}_{t}$, as follows:
\begin{equation}
    \label{eq:new_unbiased_statistics}
    \mathrm{M}_{t}=\mathrm{E}_{\mathcal{B}_{t}}[\mu_{t}], \quad
    \mathrm{\Sigma}_{t}=\frac{m}{m-1}\mathrm{E}_{\mathcal{B}_{t}}[\sigma_{t}^{2}],
\end{equation}
where $\mathcal{B}_{t}$ represents mini-batches of target training data. With these statistics estimated from the target domain, we can reformulate the batch normalization of the pre-trained model as follows:
\begin{equation}
\label{eq:reformulation_BN}
\begin{aligned}
\hat{y} & = \gamma_{s} \cdot \frac{z-\mathrm{M}_{s}}{\sqrt{\mathrm{\Sigma}_{s}+ \epsilon}} + \beta_{s} \\
        & = \gamma_{s} \cdot \frac{\sqrt{\mathrm{\Sigma}_{t}+ \epsilon}}{\sqrt{\mathrm{\Sigma}_{s}+ \epsilon}} \cdot\frac{z - \mathrm{M}_{t} +\mathrm{M}_{t} - \mathrm{M}_{s} }{\sqrt{\mathrm{\Sigma}_{t}+ \epsilon}}+ \beta_{s}, \\
        & = (\gamma_{s} \frac{\sqrt{\mathrm{\Sigma}_{t}+ \epsilon}}{\sqrt{\mathrm{\Sigma}_{s}+ \epsilon}}) \cdot\frac{z - \mathrm{M}_{t} }{\sqrt{\mathrm{\Sigma}_{t}+ \epsilon}}+ (\beta_{s}+\gamma_{s}\frac{\mathrm{M}_{t} - \mathrm{M}_{s}}{\sqrt{\mathrm{\Sigma}_{s}+ \epsilon}}) \\
        & = \gamma_{t}  \cdot\frac{z - \mathrm{M}_{t} }{\sqrt{\mathrm{\Sigma}_{t}+ \epsilon}}+ \beta_{t}, \\
\end{aligned}
\end{equation}

where $\gamma_{s}$ and $\beta_{s}$ are the learning parameters of batch normalization that are pre-trained on the source domain. According to this reformulation, we define the new batch normalization parameters, $\gamma_{t}$ and $\beta_{t}$, with the target domain statistics as follows:
\begin{equation}
    \label{eq:new_gamma_beta}
    \gamma_{t} = \gamma_{s} \cdot \frac{\sqrt{\mathrm{\Sigma}_{t}+ \epsilon}}{\sqrt{\mathrm{\Sigma}_{s}+ \epsilon}}, \quad \beta_{t} = \beta_{s}+\gamma_{s}\cdot\frac{\mathrm{M}_{t} - \mathrm{M}_{s}}{\sqrt{\mathrm{\Sigma}_{s}+ \epsilon}}.
\end{equation}

After computing these new parameters, we replace $\gamma_{s}$ and $\beta_{s}$ in the pre-trained model with $\gamma_{t}$ and $\beta_{t}$, respectively. Additionally, the statistics $\mathrm{M}_{s}$ and $\mathrm{\Sigma}_{s}$ are replaced by $\mathrm{M}_{t}$ and $\mathrm{\Sigma}_{t}$. After converting all batch normalization layers in the pre-trained model, we start training on the new target domain. With the converted BN parameters, the pre-trained model requires only slight adjustments to the batch statistics of the target domain. Furthermore, the movement of $\mathrm{M}_{t}$ and $\mathrm{\Sigma}_{t}$ can be largely reduced since the batch statistics of the target training data are already close to $\mathrm{M}_{t}$ and $\mathrm{\Sigma}_{t}$. As a result, the converted BN substantially mitigates feature distortion, leading to improved training performance on the target domain. The overall process of our BN conversion is summarized in Algorithm \ref{alg:Bn_conversion}.

\begin{algorithm}[t]
\caption{Batch Normalization Conversion}
\begin{algorithmic}
\label{alg:Bn_conversion}
\REQUIRE
Pre-trained model with BN parameters $\gamma_{s}$, $\beta_{s}$, and statistics $\mathrm{M}_{s}$, $\mathrm{\Sigma}_{s}$ from the source domain, Target training data with mini-batches $\mathcal{B}_{t}$, each containing $m$ elements;
\STATE Set pre-trained model to test mode
\FOR{each mini-batch $\mathcal{B} \in \mathcal{B}_{t}$}    
    \STATE Compute batch statistics $\mu_{t}$ and $\sigma_{t}$ with $\mathcal{B}$ for all BN layers
\ENDFOR
\STATE Compute the unbiased estimates of statistics for all BN layers: $M_{t}=\mathrm{E}_{\mathcal{B}_{t}}[\mu_{t}]$, $\Sigma_{t}=\frac{m}{m-1}\mathrm{E}_{\mathcal{B}_{t}}[\sigma_{t}^{2}]$
\STATE Compute the new parameters for all BN layers: $\gamma_{t} = \gamma_{s} \cdot \frac{\sqrt{\Sigma_{t}+ \epsilon}}{\sqrt{\Sigma_{s}+ \epsilon}}$, $\beta_{t} = \beta_{s}+\gamma_{s}\cdot\frac{M_{t} - M_{s}}{\sqrt{\Sigma_{s}+ \epsilon}}$
\STATE Replace parameters $\gamma_{s}$ with $\gamma_{t}$, and $\beta_{s}$ with $\beta_{t}$ in the pre-trained model for all BN layers\
\STATE Replace the statistics $M_{s}$ with $M_{t}$, and $\Sigma_{s}$ with $\Sigma_{t}$ in the pre-trained model for all BN layers\
\STATE Start fine-tuning on the target domain
\end{algorithmic}
\end{algorithm}

Utilizing our BN conversion technique, neural networks undergo more stable adjustments that aid in mitigating the distortion of pre-trained features, as shown in Figure \ref{fig:fmow_feature_histogram_measures}. For a comprehensive understanding of how neural network evolves during adaptation, we compute the norm of the relative change using the equation:
\begin{equation}
\label{eq:relative_change}
\frac{\|\Delta W\|_{2}}{\|W\|_{2}}=\frac{\|\widetilde{W}-W\|_{2}}{\|W\|_{2}},
\end{equation}
where $\widetilde{W}$ represents the parameter values after the training is completed, and $W$ represents the initial parameter values before training begins. To conduct our evaluation, we employ a pre-trained ResNet-50 model from MoCo-v2 \cite{chen2020improved} and perform two types of training on fMoW \cite{christie2018functional} dataset: fine-tuning without our BN conversion (FT) and fine-tuning with our BN conversion (DAFT). For each case, we compute the norm of the relative change for all layers of the ResNet-50 model. It is important to note that $W$ for FT refers to the value of the pre-trained model, whereas $W$ for DAFT refers to the value after applying BN conversion to the pre-trained model. As illustrated in Figure \ref{fig:relative_changes}, we can clearly observe that DAFT leads to significantly smaller relative changes compared to FT. This reduction is evident not only in the parameters of all layers but also in the statistics of all BN layers. This compelling evidence substantiates the effectiveness of our BN conversion technique, enabling the model to adapt more effectively to the target domain while inducing minor alterations. 

\subsection{Integrating LP and FT}
Our BN conversion technique can be applied to various transfer learning algorithms for neural networks that incorporate BN layers. However, considering the limitations of the recent LP-FT method, we integrate LP and FT in a single stage to leverage BN conversion more effectively.
\subsubsection{Limitation of LP-FT}
Let $\theta$ represent the parameters of the feature extractor denoted as $f_{\theta}$, initialized with a pre-trained model. The classifier with parameters $w$ is denoted as $g_{w}$. While it is commonly a linear layer, it could also consist of multiple layers. In the LP-FT approach, the head classifier $g_{w}$ is initialized using the value optimized by LP, and then FT is performed with a very small learning rate $\eta$. As a result, the head layer undergoes only minor changes during FT and remains nearly unchanged after the FT stage. This implies that the head layer optimization primarily occurs during the LP stage, and it is not optimized in conjunction with the feature extractor's adaptation to the target domain. This limitation hinders performance, and it becomes particularly critical when there is a significant difference in data distribution between the source and target domains. To address this issue, we introduce integrating LP and FT into a single stage, allowing the head layer to be optimized with gradually adapting features.

\subsubsection{Integrated LP-FT} In order to integrate the LP and FT stages, we introduce separate learning rates, $\eta_{\theta}$ and $\eta_{w}$, for the feature extractor $f_{\theta}$ and the head layer $g_{w}$. While the conventional FT also suggests using a different learning rate for the head layer that is 10 times larger than that of the feature extractor, we independently determine the optimized learning rates for each component. To efficiently determine these optimized learning rates, we adopt a two-step procedure, sweeping over each learning rate. In the first step, we sweep over the learning rates $\eta_{\theta}$ for the feature extractor with fixing $\eta_{w}$. Having determined the optimized learning rate $\eta_{\theta}$ for the feature extractor, we move on to the second step, where we sweep over the learning rates $\eta_{w}$ while utilizing the previously determined optimized learning rate $\eta_{\theta}$. By following this process, we integrate the FT and LP stages into a single approach, which distinguishes our approach from LP-FT. Moreover, this strategy significantly reduces the number of hyperparameter combinations and streamlines the search for proper learning rates.

Additionally, we employ a zero initialization strategy for the head layer. This approach is aimed at reducing the influence of gradients on the feature extractor during the initial training phases. By initializing the head layer with zeros, we try to avoid making significant changes to the feature extractor until the head layer has started to converge to a reasonable solution. However, in cases where the head layer comprises multiple layers, we opt for random initialization instead of zero values to enable gradient computation. Detailed distinctions between our DAFT and other fine-tuning techniques, including FT and LP-FT, are summarized in Table \ref{table:summary}.

\begin{table}[t]
    \renewcommand{\arraystretch}{1.2}
    \renewcommand{\tabcolsep}{1.0mm}
    \centering
    \begin{tabular}{lcccc}
    \toprule
    Method & Head Init  &  LR for FT & BN mode & BN Conversion\\
    \midrule \midrule
    FT & Random & $\eta_{\theta}=\eta_{w}$ & train & X \\
    LP-FT & LP & $\eta_{\theta}=\eta_{w}$ & test & X  \\
    DAFT & Zero & $\eta_{\theta} \neq \eta_{w}$ & train & O \\
    \bottomrule
    \end{tabular}
    \caption{Summary of details for our DAFT and other fine-tuning methods.}
    \label{table:summary}
\end{table}

\section{Experiments}
In this section, we present the results of various experiments conducted to evaluate the effectiveness of our proposed method. We first present the experimental results on classification tasks and segmentation tasks. Additionally, we provide the results of additional experiments, including robustness evaluations and ablation studies.

\subsection{Experiments on Classification Task}

\begin{table*}[t]
\begin{center}
\begin{tabular}{llccccccccccc}
\toprule
\multirow{2}{*}{Model} & \multirow{2}{*}{Method} & \multicolumn{3}{c}{CIFAR-10} & \multicolumn{2}{c}{Entity-30} &   \multicolumn{2}{c}{Living-17} &   \multicolumn{2}{c}{DomainNet}&   \multicolumn{2}{c}{fMoW} \\
\cmidrule(lr){3-5} \cmidrule(lr){6-7} \cmidrule(lr){8-9} \cmidrule(lr){10-11} \cmidrule{12-13} 
& & ID & CIFAR-10.1 & STL & ID & OOD & ID & OOD & ID & OOD & ID & OOD \\
\midrule \midrule
\multirow{4}{*}{MoCo}
& LP    & 91.78 & 82.53 & 86.01 & 91.35 & 64.10 & 96.27 & 82.06 & 73.67 & 64.41 & 49.12 & 34.92 \\
& FT    & 97.66 & 92.95 & 81.18 & 93.39 & 62.48 & 96.98 & 78.41 & 87.77 & 62.19 & 60.72 & 41.04 \\
& LP-FT & 97.66 & 93.60 & 90.79 & 93.80 & 62.03 & 97.55 & 82.53 & 85.56 & 70.89 & 55.24 & 38.50 \\
& DAFT  & \textbf{97.84} & \textbf{94.18} & \textbf{92.36} & \textbf{94.54} & \textbf{65.11} & \textbf{97.84} & \textbf{84.35} & \textbf{89.31} & \textbf{74.67} & \textbf{61.91} & \textbf{43.29} \\
\midrule
\multirow{4}{*}{CLIP}
& LP    & 87.87 & 77.62 & 84.82 & 91.11 & 63.99 & 94.61 & \textbf{79.57} & 89.13 & 82.29 & 49.85 & 31.24 \\
& FT    & 93.88 & 87.07 & 69.50 & 93.74 & 60.28 & 94.69 & 66.96 & 84.67 & 53.91 & 54.83 & 31.65 \\
& LP-FT & 94.02 & 87.77 & 87.01 & 93.29 & 62.62 & 95.73 & 77.96 & 91.27 & 82.53 & 60.73 & 39.93 \\
& DAFT  & \textbf{95.49} & \textbf{89.52} & \textbf{87.51} & \textbf{94.40} & \textbf{64.34} & \textbf{96.47} & 78.24 & \textbf{91.72} & \textbf{82.57} & \textbf{64.65} & \textbf{43.40}\\
\midrule
\multirow{4}{*}{SWAV}
& LP    & 93.24 & 85.23 & 89.58 & 92.05 & 63.91 & 96.67 & 80.06 & 75.13 & 59.71 & 47.29 & 29.74 \\
& FT    & 97.55 & 93.25 & 88.26 & 94.96 & 63.03 & 97.61 & 80.65 & 88.34 & 66.69 & 66.06 & 45.67 \\
& LP-FT & 97.50 & 92.60 & \textbf{91.44} & 94.67 & 61.78 & 97.47 & 80.84 & 85.19 & 67.38 & 58.27 & 37.62 \\
& DAFT  & \textbf{97.77} & \textbf{93.82} & 89.96 & \textbf{95.07} & \textbf{63.96} & \textbf{97.88} & \textbf{84.06} & \textbf{88.80} & \textbf{69.03} & \textbf{66.93} & \textbf{46.14} \\
\bottomrule
\end{tabular}%
\end{center}
\caption{
ID and OOD accuracies (\%) of three different pretrained models on five different dataset.  The upper part of the first row represents the training data and the lower part represents the test data. CIFAR-10.1 and STL are used as OOD datasets for CIFAR-10, as there is no dedicated OOD dataset for CIFAR-10. Best results are highlighted in bold. 
}
\label{table:classification_task}
\end{table*}

\subsubsection{Dataset}
We conduct experiments on various classification tasks and evaluate both In-Distribution (ID) and Out-of-Distribution (OOD) accuracy using the following datasets.

\begin{itemize}

    \item \textbf{CIFAR-10} \cite{krizhevsky2009learning}: A dataset that contains 10 categories of objects. For the OOD dataset, we use two additional datasets, \textbf{CIFAR-10.1} \cite{recht2018cifar} and \textbf{STL} \cite{coates2011analysis}. CIFAR-10.1 is a subset of the TinyImage dataset \cite{birhane2021large} and has the same categories as CIFAR-10. STL has the same categories as CIFAR-10 except for `monkey' instead of `frog'. Therefore, we remove the `monkey' class in STL to align with CIFAR-10 categories. For data augmentation, we resize the input image to 224x224 and apply random horizontal flip.
    
    \item \textbf{Entity-30} and \textbf{Living-17} \cite{santurkar2020breeds}: Part of the BREEDS benchmarks, which are subpopulation shift datasets constructed using ImageNet. Entity-30 contains 30 categories of objects, and Living-17 contains 17 categories of animals. Each dataset consists of source and target domains. For both Entity-30 and Living-17, we use the source domain as the ID dataset and the target domain as the OOD dataset. Data augmentation involves RandomResizedCrop to 224x224 and random horizontal flip.
    
    \item \textbf{DomainNet} \cite{peng2019moment}: A dataset that includes 345 categories of common objects in six different domain including Clipart, Infograph, Painting, Quickdraw, Real, and Sketch. However, due to labeling noise in DomainNet, we used a subset of DomainNet \cite{tan2020class} containing 40 categories of common objects from the Sketch, Real, Clipart, and Painting domains. Sketch domain is used for the ID dataset and the rest of the domains (Real, Clipart, Painting) are used for the OOD dataset. Data augmentation includes resizing the images to 256x256, random cropping to 224x224, and applying random horizontal flip.
    
    \item \textbf{fMoW} \cite{christie2018functional}: A remote sensing dataset that contains 62 categories of satellite images in five different regions including Asia, Europe, Africa, Americas, and Oceania. For our evaluation, we use images from the Americas region as the ID dataset, and images from the Europe and Africa regions are used for the OOD dataset. As the fMoW data size is 224x224, we apply only random horizontal flip for data augmentation.
\end{itemize}

\subsubsection{Pretrained Models}
To ensure the generalizability of our method across different pre-trained models, we utilize three different models based on the ResNet-50 architecture: MoCo-v2 \cite{chen2020improved}, CLIP \cite{radford2021learning}, and SWAV \cite{caron2020unsupervised}. Exceptionally, for the fMoW dataset, we use MoCo-TP \cite{ayush2021geography} instead of MoCo-v2 as the pre-trained model.

\subsubsection{Training Details} 
Consistent with LP-FT \cite{kumar2022fine}, we use $\ell_2$-regularized logistic regression classifier for linear probing and also choose the best $\ell_2$-regularization hyperparameter based on ID validation accuracy. For fine-tuning, we employ an SGD classifier with a cosine learning rate schedule and a batch size of 64. We fine-tune for 20 epochs on CIFAR-10, Entity-30, and Living-17, but extend the fine-tuning to 50 epochs on DomainNet and fMoW due to their limited number of images. Early stopping is applied based on the ID validation accuracy.

\subsubsection{Results} Our experimental results on five distinct datasets are summarized in Table \ref{table:classification_task}. All of results are averaged over three separate runs. First, we observe that FT tends to exhibit superior performance in ID accuracy compared to LP, while LP tends to outperform FT in OOD accuracy. Across three diverse pre-trained models, our DAFT consistently achieves higher ID accuracy on all datasets in comparison to other baseline methods. Additionally, DAFT consistently exhibits higher OOD accuracy across most of the datasets.

\begin{table}[t]
    \centering
    \begin{tabular}{lcc}
    \toprule
    Method & VOC & Cityscapes\\
    \midrule \midrule
    LP      & 75.85  & 65.69 \\
    FT      & 78.47 & 74.42 \\
    LP-FT   & 77.13 & 66.41 \\
    DAFT    & \textbf{79.23} & \textbf{76.03} \\
    \bottomrule
    \end{tabular}
    \caption{Mean Intersection over Union (\%) for segmentation datasets. Best results are highlighted in bold.}
    \label{table:segmentation}
\end{table}

\begin{table*}[t]
\begin{center}
\renewcommand{\arraystretch}{1.0}
\renewcommand{\tabcolsep}{0.6mm}

\begin{tabular}{llcccccccccccccccc}
\toprule
\multirow{2}{*}{Model} & \multirow{2}{*}{Method} & \multicolumn{3}{c}{Noise} & \multicolumn{4}{c}{Blur} &   \multicolumn{4}{c}{Weather} &   \multicolumn{4}{c}{Digital} &  \multirow{2}{*}{mCE} \\
\cmidrule(lr){3-5} \cmidrule(lr){6-9} \cmidrule(lr){10-13} \cmidrule{14-17} 
 & & \multicolumn{1}{c}{Gauss.} & Shot & Impulse & Defocus & Glass & Motion & Zoom & Snow & Frost & Fog & Bright & Cont. & Elastic & Pixel & JPEG &  \\ 
\midrule \midrule
\multirow{4}{*}{MoCo} 
& LP   & 47.5 & 40.1 & 54.2 & 19.5 & 50.8 & 33.4 & 23.5 & 22.8 & 23.6 & 23.7 & 10.1 & 18.3 & 28.7 & 30.5 & 25.7 & 30.2 \\
& FT & 56.1 & 41.9 & \textbf{45.5} & 9.9 & 41.0 & 17.1 & 12.4 & 9.3 & 11.9 & 7.4 & 3.2 & 9.8 & \textbf{12.6} & 27.2 & 20.9 & 21.7 \\
& LP-FT & 46.5 & 36.0 & 48.3 & 8.1 & \textbf{38.2} & \textbf{14.6} & \textbf{10.4} & \textbf{9.1} & 11.3 & \textbf{7.1} & 3.3 & 11.0 & 12.9 & \textbf{24.6} &\textbf{19.2} & 20.0 \\
& DAFT & \textbf{37.4} & \textbf{28.8} & 46.7 & \textbf{8.0} & 39.8 & 16.3 & 10.8 & 9.3 & \textbf{10.9} & 8.2 & \textbf{3.2} & \textbf{9.1} & 14.4 & 28.5 & 19.5 & \textbf{19.4}\\
\midrule
\multirow{4}{*}{CLIP} 
& LP    & 65.1 & 57.7 & 56.2 & 24.5 & 57.4 & 35.3 & 27.8 & 26.5 & 27.9 & 23.4 & 15.4 & 31.8 & 30.1 & 28.5 & 33.8 & 36.1 \\
& FT    & \textbf{46.5} & \textbf{35.1} & \textbf{37.2} & 17.6 & 52.2 & 24.8 & 22.0 & 19.6 & 22.2 & 12.7 & 8.4 & 24.7 & 19.0 & \textbf{27.6} & \textbf{18.7} & 25.9 \\
& LP-FT & 57.3 & 46.1 & 48.5 & 14.4 & 51.6 & 24.4 & 18.3 & 15.8 & 18.2 & 11.6 & 8.1 & 23.8 & 20.4 & 33.9 & 27.3 & 28.0 \\
& DAFT & 61.6 & 48.1 & 40.7 & \textbf{13.6} & \textbf{47.0} & \textbf{22.0} & \textbf{16.9} & \textbf{13.8} & \textbf{15.9} & \textbf{9.7} & \textbf{6.2} & \textbf{18.4} & \textbf{17.6} & 32.0 & 23.0 & \textbf{25.8} \\

\midrule
\multirow{4}{*}{SWAV} 
& LP    & 50.0 & 42.8 & 60.2 & 17.4 & 45.8 & 27.2 & 20.2 & 19.3 & 22.3 & 19.6 & 8.6 & 12.5 & 25.0 & 27.8 & 25.2 & 28.3 \\
& FT    & 51.7 & 38.6 & 46.9 & 10.1 & 39.1 & 16.8 & 11.5 & \textbf{8.7} & 10.6 & 7.7 & 3.6 & 10.9 & \textbf{12.4} & 29.2 & 21.0 & 21.3 \\
& LP-FT & 50.1 & 37.7 & \textbf{36.6} & \textbf{8.7} & \textbf{34.0} & \textbf{15.4} & \textbf{10.4} & 8.8 & 11.1 & \textbf{6.6} & 3.4 & 9.4 & 12.5 & \textbf{25.3} & 20.5 & 19.4 \\
& DAFT  & \textbf{41.4} & \textbf{30.2} & 46.9 & 8.9 & 36.1 & 16.0 & 10.9 & 8.9 & \textbf{10.0} & 7.6 & \textbf{3.1} & \textbf{7.7} & 12.8 & 25.4 & \textbf{19.5} & \textbf{19.0} \\
\bottomrule
\end{tabular}%

\end{center}
\caption{
Corruption Error (\%) for 15 types of corruptions. The mean Corruption Error (mCE) is calculated for all corruptions. A lower value of Corruption Error indicates better performance. Best results are highlighted in bold.
}
\label{table:cifar10_robustness}
\end{table*}

\subsection{Experiments on Segmentation Task}
\subsubsection{Dataset}
For the segmentation task, we use the following datasets:
\begin{itemize}
    \item \textbf{Pascal VOC2012} \cite{everingham2015pascal}: This dataset consists of 21 classes for semantic segmentation, including 20 object classes (e.g., person, car, bicycle, etc.) and an additional background class. To augment the data, we apply random scaling between 0.5 and 2.0, random crop of size 513x513, and random horizontal flip. Additionally, we use the augmented dataset with extra annotations \cite{abadi2016tensorflow}.
    \item \textbf{Cityscapes} \cite{cordts2016cityscapes}: This dataset includes 19 classes for semantic segmentation, comprising 18 object classes (e.g., cars, pedestrians, buildings, roads, etc.) and an additional background class. To augment the data, we use random crop of size 768x768, color jitter with brightness 0.5, contrast 0.5, saturation 0.5, and hue 0, and random horizontal flip.
\end{itemize}

\subsubsection{Training Details}
For the segmentation task, we adopt the DeepLabv3+ \cite{chen2018encoder} framework and initialize its backbone with a pre-trained ResNet-50 model from MoCo-v2. During fine-tuning, we use the SGD optimizer with a batch size of 16 and a polynomial learning rate schedule with power 0.9. We conduct the fine-tuning process for a total of 30,000 iterations, and the final model is evaluated thereafter.

\subsubsection{Results} Table \ref{table:segmentation} presents the results of our experiments on the segmentation task. Our DAFT demonstrates improved performance compared to other baseline methods for both the VOC and Cityscapes datasets. LP-FT also shows some enhancement by employing an LP-trained model, but it still achieves inferior performance compared to FT. These results show the significance of BN in the context of the segmentation task as well.

\begin{table}[t]
    \centering
    \resizebox{0.99\columnwidth}{!}{%
    \begin{tabular}{llccc}
    \toprule
    \multirow{2}{*}{Model} & \multirow{2}{*}{Method} & \multicolumn{3}{c}{CIFAR-10} \\
    \cmidrule{3-5}
    & & ID & CIFAR-10.1 & STL\\
    \midrule \midrule
    \multirow{6}{*}{MoCo}
    & FT                & 97.66 & 92.95 & 81.18 \\
    & + BN conversion   & 97.59 & 93.27 & 84.05 \\   
    \cmidrule{2-5}
    & LP-FT             & 97.66 & 93.60 & 90.79 \\
    & + BN conversion   & 97.31 & 93.00 & 91.66 \\     
    \cmidrule{2-5}
    & Integrated LP-FT & 97.63 & 94.12 & 90.52 \\
    & + BN conversion       & 97.84 & 94.18 & 92.36 \\     
    \midrule
    \multirow{6}{*}{CLIP}
    & FT                & 93.88 & 87.07 & 69.50 \\
    & + BN conversion   & 95.40 & 88.8 & 81.47 \\
    \cmidrule{2-5}
    & LP-FT             & 94.02 & 87.77 & 87.01 \\
    & + BN conversion   & 94.70 & 88.50 & 85.49 \\
    \cmidrule{2-5}
    & Integrated LP-FT & 94.29 & 87.32 & 73.12 \\
    & + BN conversion  & 95.49 & 89.52 & 87.51 \\
    \midrule
    \multirow{6}{*}{SWAV}
    & FT                & 97.55  & 93.25 & 88.26 \\
    & + BN conversion   & 97.62  & 93.62 & 89.00\\
    \cmidrule{2-5}
    & LP-FT             & 97.50 & 92.60 & 91.44 \\
    & + BN conversion   & 97.61 & 93.45 & 89.35 \\  
    \cmidrule{2-5}
    & Integrated LP-FT     & 97.28 & 92.53 & 89.46 \\
    & + BN conversion   & 97.77 & 93.82 & 89.96 \\ 
    \bottomrule
    \end{tabular}
    }
    \caption{BN conversion effect for each fine-tuning method.}
    \label{table:ablation}
\end{table}

\subsection{Additional Experiments}

\subsubsection{Robustness}
In addition to ID and OOD test, we also evaluate the robustness of our method. After fine-tuning on the CIFAR-10 dataset, we evaluate the model's performance on the CIFAR-10-C dataset \cite{hendrycks2019benchmarking}. The CIFAR-10-C dataset comprises 15 types of corruptions grouped into four main categories: noise, blur, weather, and digital. Each corruption type is present at five different levels of severity. We calculate the Corruption Error for all levels of corruptions and then compute their average. The results of the robustness assessment are presented in Table \ref{table:cifar10_robustness}. Interestingly, LP outperforms FT in terms of OOD accuracy, but it shows weaker performance in terms of robustness. And, our DAFT achieves lower average errors across various corruptions compared to other methods. Since this robustness experiment is conducted for the methods whose results are presented in Table \ref{table:classification_task}, the Corruption Error values are also obtained as an average over three runs.

\subsubsection{Ablation Study}

Table \ref{table:ablation} presents the results of our ablation study on BN conversion and integrated LP-FT. We conducted the study using a pre-trained ResNet-50 model from MoCo-v2 \cite{chen2020improved} on the CIFAR-10 dataset. To evaluate the effectiveness of BN conversion alone, we applied it to each fine-tuning method, including FT, LP-FT, and integrated LP-FT. The results demonstrate that BN conversion is effective for most of OOD tests with all fine-tuning methods, indicating its importance in improving performance on OOD datasets. Additionally, we observe that integrated LP-FT without BN conversion performs similarly to LP-FT, revealing the limitations of the two-stage optimization and emphasizing the critical role of BN in the overall optimization process.

\section{Conclusion}
In this paper, we introduce Domain-Aware Fine-Tuning (DAFT), a novel approach designed to enhance the adaptability and performance of fine-tuned neural networks. Our method optimizes performance and minimizes network modification by aligning batch normalization layers with the target domain. Additionally, the integration of LP and FT allows the head layer to be optimized with gradually adapting features. The widespread use of batch normalization layers in many practical networks makes DAFT a valuable solution for real-world applications. Overall, DAFT bridges the gap between pre-trained models and new target domains, which contributes to improved model performance and generalization.

\section{Acknowledgments}
This work was supported in part by National Research Foundation of Korea (NRF,
2021R1A2C2014504(20\%) and 2021M3F3A2A02037893(20\%)), in part by Institute of Information \& communications Technology Planning \& Evaluation (IITP) grant funded by the Ministry of Science and ICT (MSIT) (2021-0-00106(15\%), 2021-0-02068(15\%), 2021-0-01080(15\%), and 2021-0-01341(15\%)), and in part by AI Graduate School Program, CAU, INMAC and BK21 FOUR program.


\clearpage
\appendix

\section{Supplementary Material}
Our code is available at \url{https://github.com/skhnha/DAFT}

\subsection{Hyperparameter Details}
\subsubsection{FT}
Regarding the hyperparameter configuration of learning rates $\eta$ for FT, we conduct a logarithmic sweep over learning rates, ranging over values such as $0.3, 0.1, \ldots, 3\times10^{-4}, 1\times10^{-4}$.

\subsubsection{LP-FT}
For LP-FT, we similarly perform a logarithmic sweep over learning rates $\eta$, ranging over values like $1 \times 10^{-4}, 3\times10^{-5}, \ldots, 1 \times 10^{-6}, 3 \times 10^{-7}$ for both MoCo and CLIP models. Additionally, we explore logarithmic learning rates $\eta$ ranging over values like $0.01, 0.003, \ldots, 1 \times 10^{-5}, 3 \times 10^{-6}$ for the SWAV model.

\subsubsection{DAFT}
For our DAFT approach, we fist perform a sweep over learning rates $\eta_{\theta}$ for the feature extractor, ranging over values like $0.1, 0.03, \ldots, 1 \times 10^{-5}, 3 \times 10^{-6}$ while fixing the learning rate $\eta_{w}$ for the linear head at 1.0. Subsequently, after identifying the optimized $\eta_{\theta}$, we sweep over learning rates $\eta_{w}$ for the linear head, ranging over values like $10, 3, 0.3, 0.1$.

\begin{table}[h]
    \centering
    \begin{tabular}{llcc}
    \toprule
        Model & Method &  ID accuracy & OOD accuracy \\
    \midrule \midrule
        \multirow{4}{*}{MoCo}
        & LP       &  80.44 & 69.01 \\
        &   FT     &  87.30 & 69.71 \\
        &   LP-FT  &  85.96 & 73.06 \\
        &   DAFT   &  \textbf{88.29} & \textbf{75.66} \\
        \midrule
        \multirow{4}{*}{CLIP}
        & LP       &  82.51 & 69.92 \\
        &   FT     &  84.36 & 61.56 \\
        &   LP-FT  &  87.01 & 72.97 \\
        &   DAFT   &  \textbf{88.55} & \textbf{74.26} \\
        \midrule
        \multirow{4}{*}{SWAV}
        & LP       &  80.88 & 68.04 \\
        &   FT     &  88.90 & 72.93 \\
        &   LP-FT  &  86.62 & 71.94 \\
        &   DAFT   &  \textbf{89.29} & \textbf{74.50} \\
    \bottomrule
    \end{tabular}
    \caption{Average accuracies (\%) of three different pre-trained models across five different tasks. The values are the average accuracies obtained from the results in Table \ref{table:classification_task}. Our DAFT outperforms other methods across three different pre-trained models in both ID and OOD tests.}
    \label{tab:summary_transfer_learning}
\end{table}

\begin{figure}[ht]
    \centering
    \resizebox{0.99\columnwidth}{!}{%
    \includegraphics{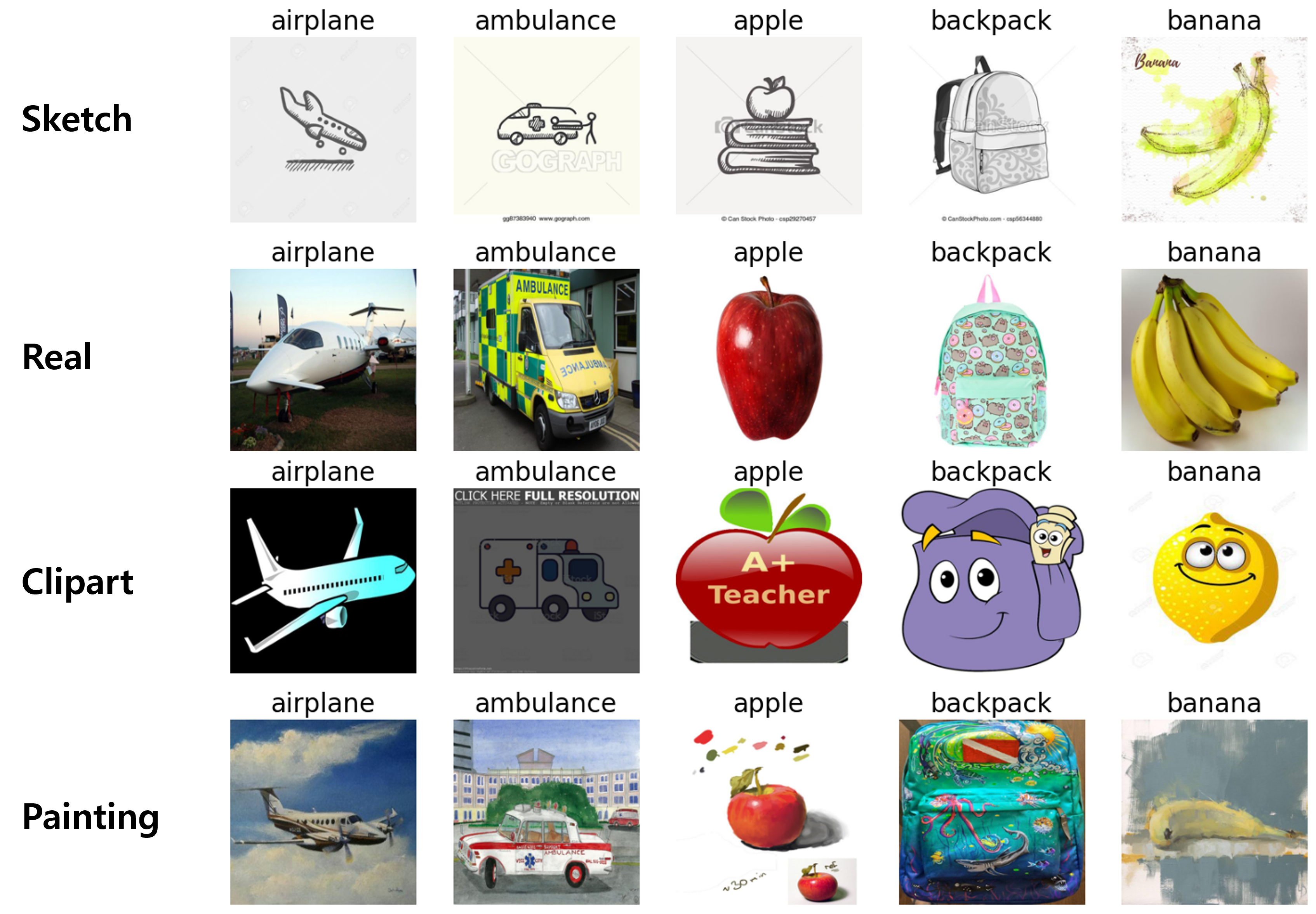}
    }
    \caption{Examples of DomainNet dataset.}
    \label{fig:DomainNet_samples}
\end{figure}

\begin{figure}[ht]
    \centering
    \resizebox{0.99\columnwidth}{!}{\includegraphics{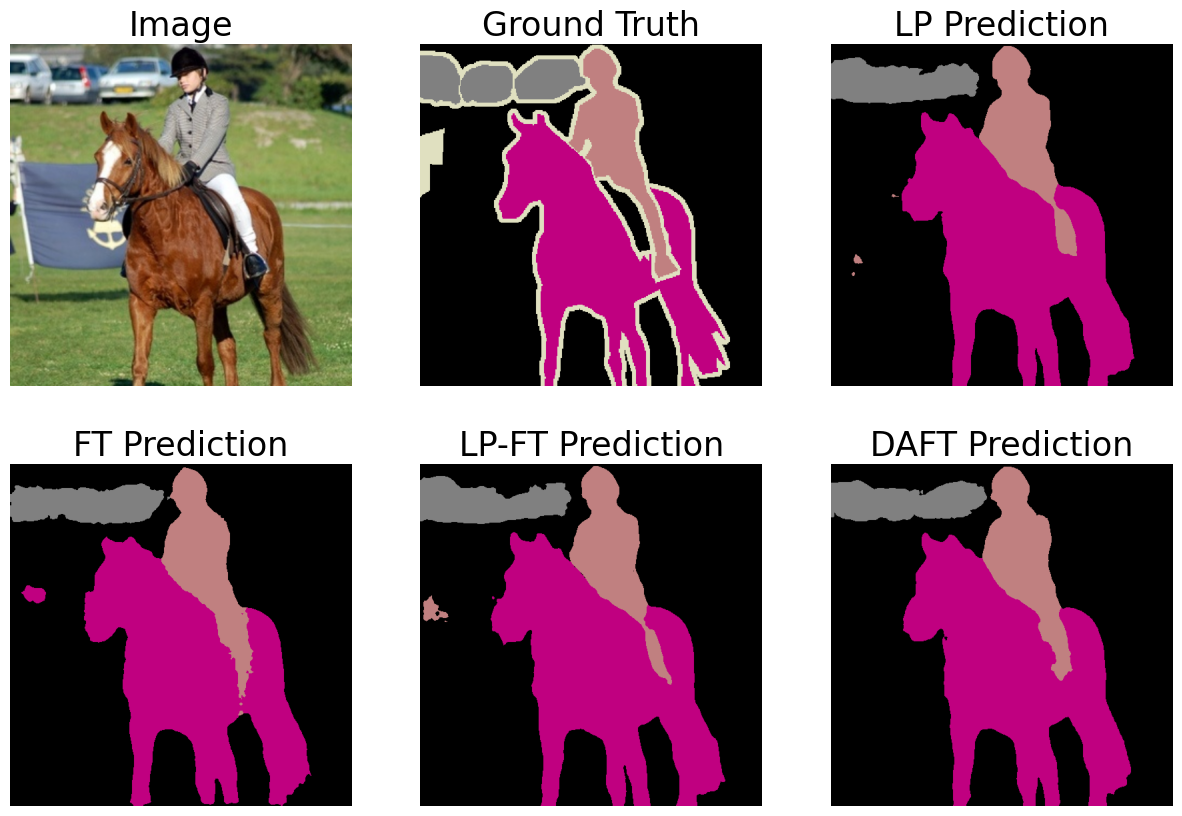}
    }
    \caption{Example of Pascal VOC 2012.}
    \label{fig:voc}
\end{figure}

\begin{figure}[ht]
    \centering
    \resizebox{0.99\columnwidth}{!}{    \includegraphics{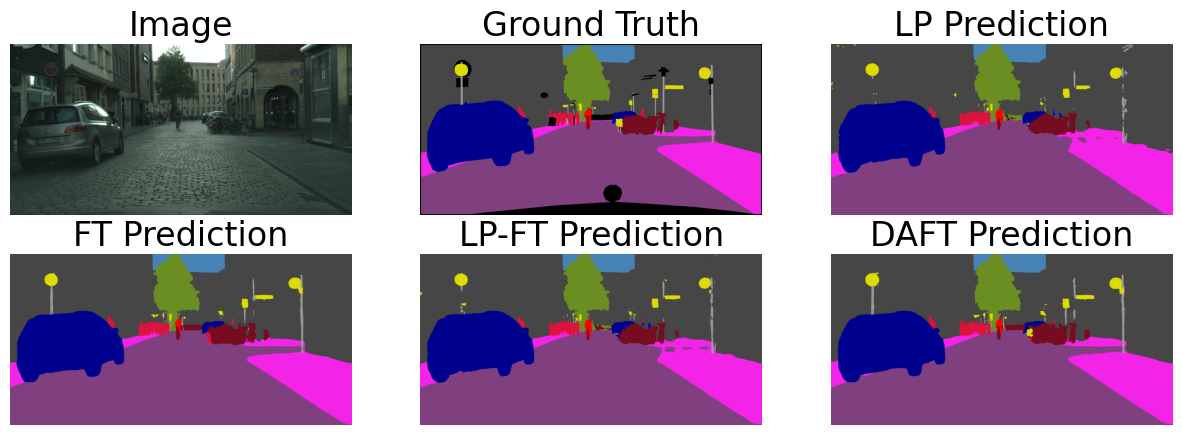}
    }
    \caption{Example of cityscapes.}
    \label{fig:cityscapes}
\end{figure}

\clearpage

\subsection{Visualizing with Cosine Similarity}
\begin{figure}[ht]
    \centering
        \begin{subfigure}[b]{0.99\columnwidth}         
         \includegraphics[width=\columnwidth]{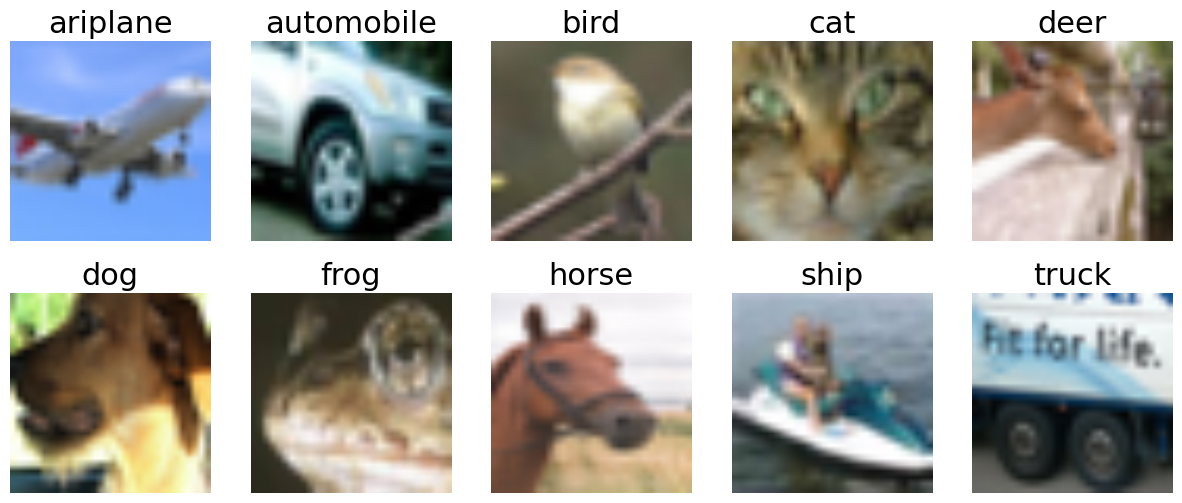}
         \caption{CIFAR-10 test images}
        \label{fig:cifar10-cos_high-test-images}
        \end{subfigure}
    \medskip
        \begin{subfigure}[b]{0.99\columnwidth}         
         \includegraphics[width=\columnwidth]{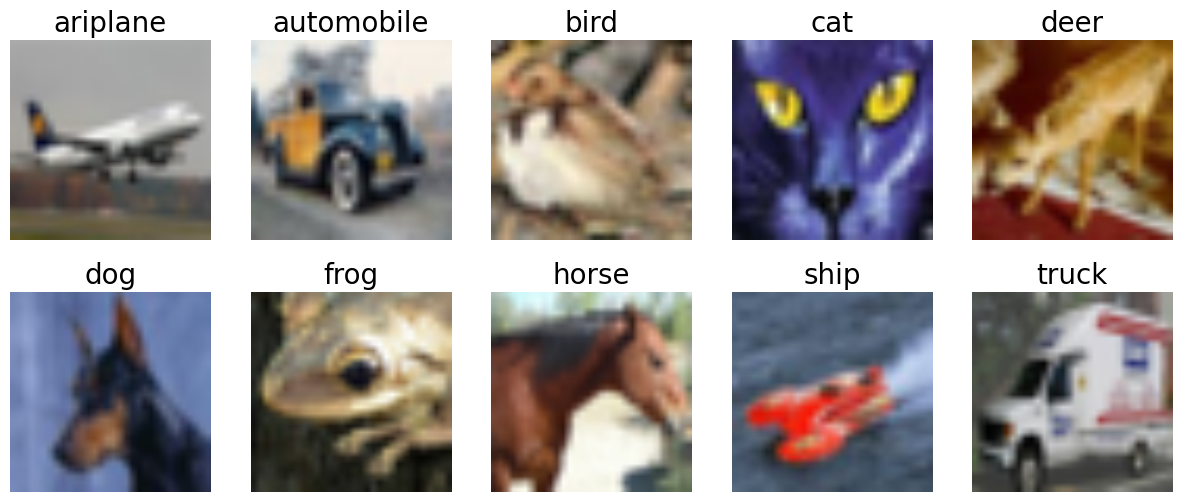}
         \caption{CIFAR-10.1 test images}
        \label{fig:ciafr10.1-cos_high-test-images}
        \end{subfigure}
    \medskip     
        \begin{subfigure}[b]{0.99\columnwidth}         
         \includegraphics[width=\columnwidth]{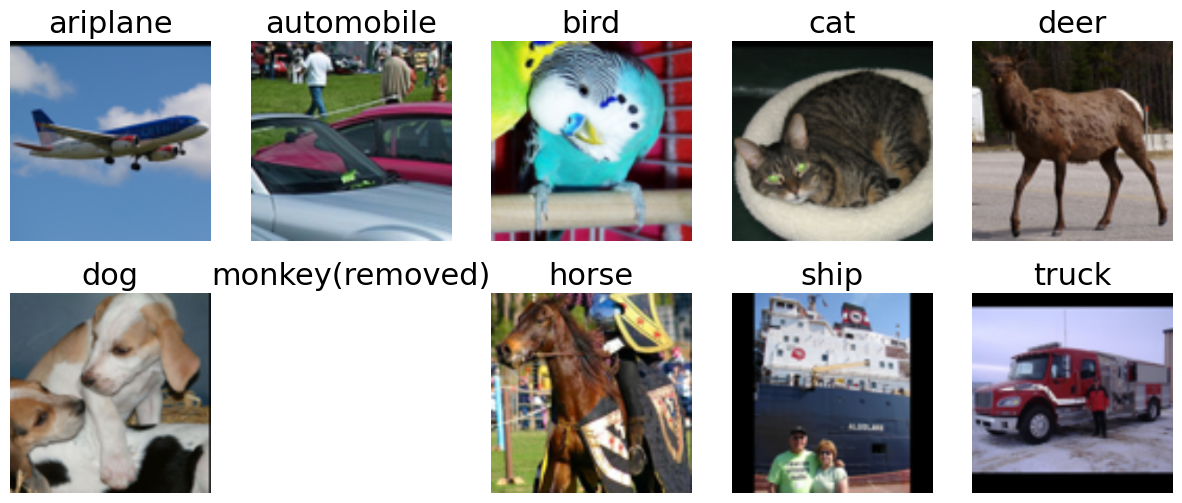}
         \caption{STL test images}
        \label{fig:STL-cos_high-test-images}
        \end{subfigure}    
    \caption{Images with the highest cosine similarity of features between before and after FT for each class.
    }
    \label{fig:cos_high-test-images}
\end{figure}

\begin{figure}[ht]
    \centering
        \begin{subfigure}[b]{0.99\columnwidth}         
         \includegraphics[width=\columnwidth]{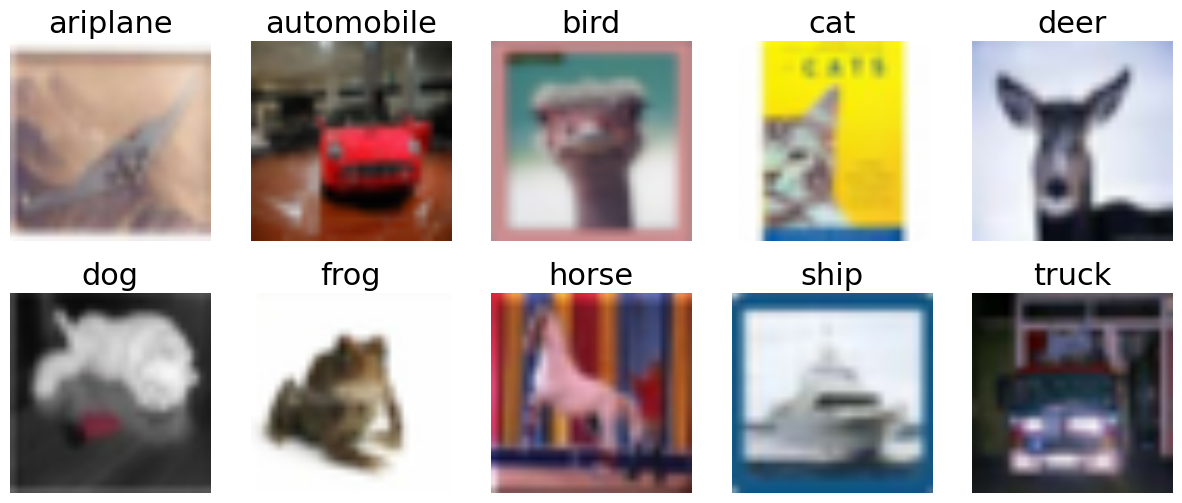}
         \caption{CIFAR-10 test images}
        \label{fig:cifar10-cos_low-test-images}
        \end{subfigure}
    \medskip
        \begin{subfigure}[b]{0.99\columnwidth}         
         \includegraphics[width=\columnwidth]{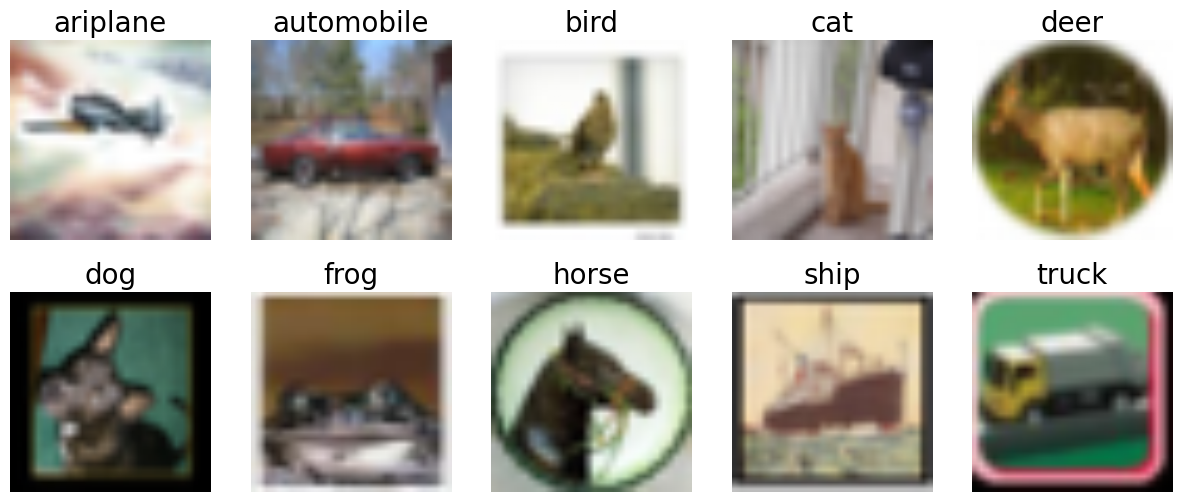}
         \caption{CIFAR-10.1 test images}
        \label{fig:ciafr10.1-cos_low-test-images}
        \end{subfigure}
    \medskip     
        \begin{subfigure}[b]{0.99\columnwidth}         
         \includegraphics[width=\columnwidth]{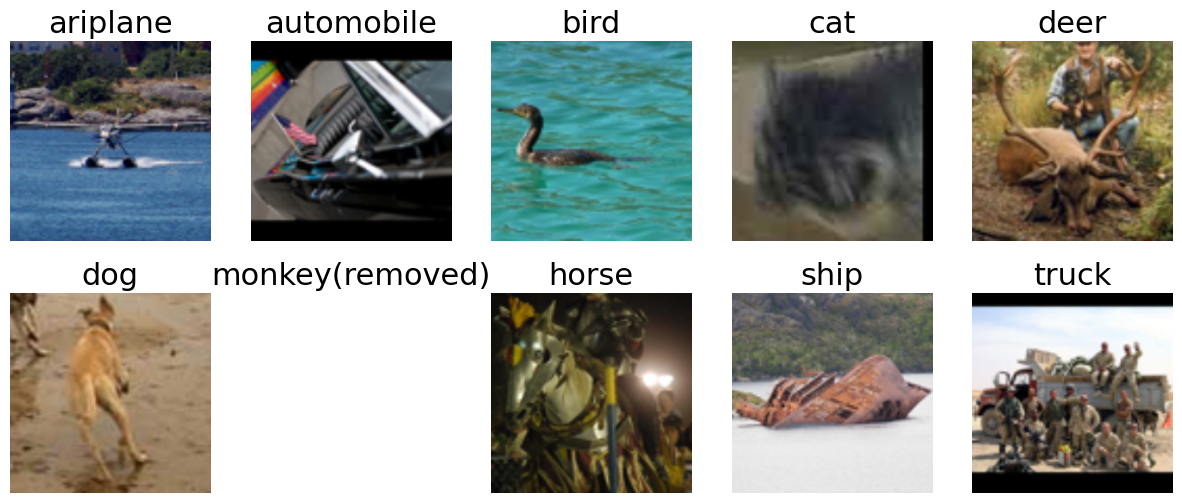}
         \caption{STL test images}
        \label{fig:STL-cos_low-test-images}
        \end{subfigure}    
    \caption{Images with the lowest cosine similarity of features between before and after FT for each class.
    }
    \label{fig:cos_low-test-images}
\end{figure}

In this section, we utilize visualizations to gain a deeper understanding of cosine similarity. Specifically, for the fine-tuning (FT) process on the CIFAR-10 dataset, we compute the cosine similarity between features before and after FT for images from CIFAR-10, CIFAR-10.1, and STL. We then visualize the test images with the highest and lowest cosine similarity for each class. Figure \ref{fig:cos_high-test-images} presents images with the highest cosine similarity, while Figure \ref{fig:cos_low-test-images} presents images with the lowest cosine similarity. Interestingly, images with the lowest cosine similarity often exhibit more spurious features, such as unusual background elements, compared to images with the highest cosine similarity. This observation suggests that these spurious features contribute to feature distortion. Consequently, it becomes crucial to mitigate the impact of such features on the feature extractor during the fine-tuning process.

\begin{figure}[htbp]
    \centering
        \begin{subfigure}[b]{0.99\columnwidth}         
         \includegraphics[width=\columnwidth]{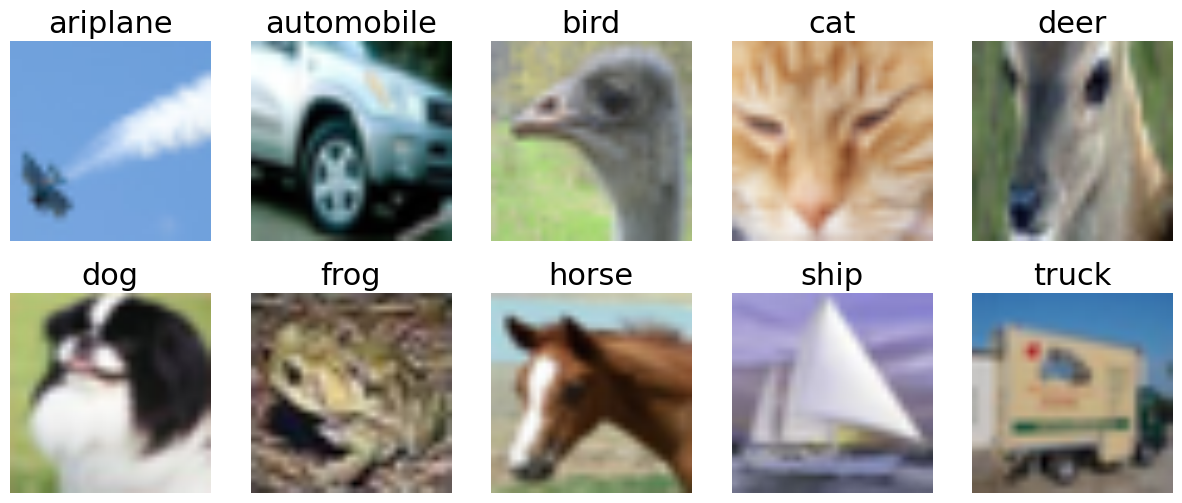}
         \caption{CIFAR-10 test images}
        \label{fig:cifar10-cos_high-test-images-daft}
        \end{subfigure}
    \medskip
        \begin{subfigure}[b]{0.99\columnwidth}         
         \includegraphics[width=\columnwidth]{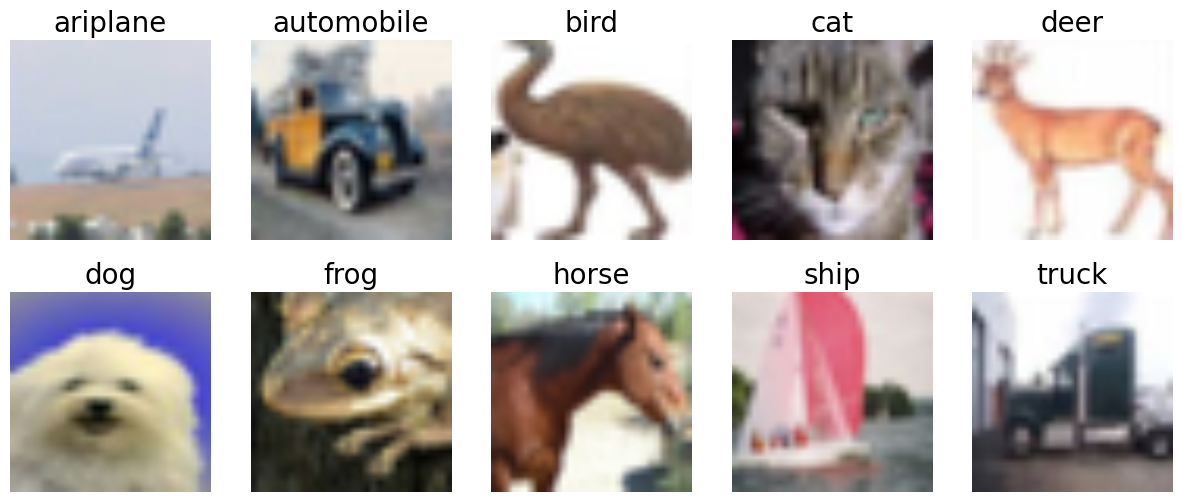}
         \caption{CIFAR-10.1 test images}
        \label{fig:ciafr10.1-cos_high-test-images-daft}
        \end{subfigure}
    \medskip     
        \begin{subfigure}[b]{0.99\columnwidth}         
         \includegraphics[width=\columnwidth]{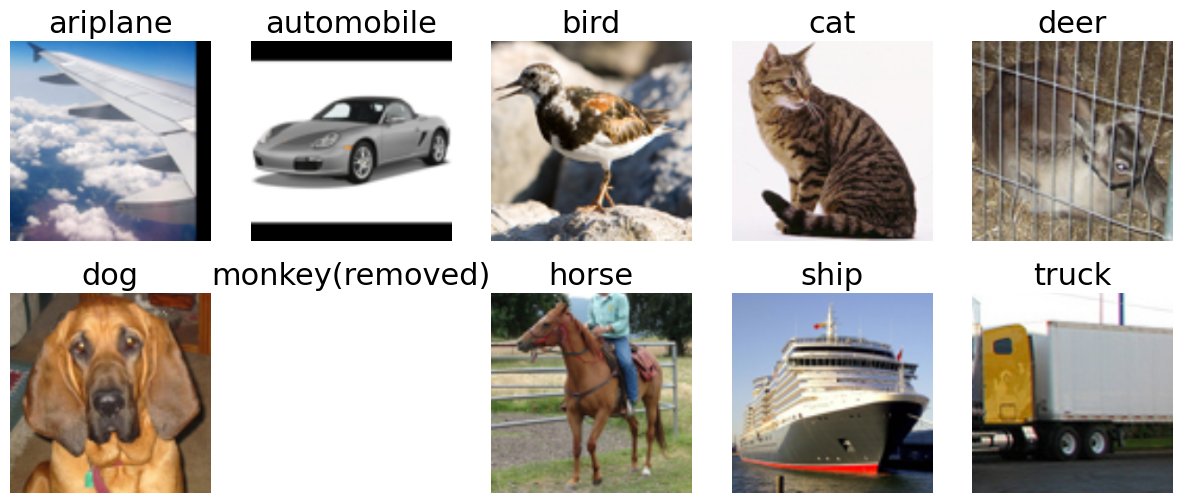}
         \caption{STL test images}
        \label{fig:STL-cos_high-test-images-daft}
        \end{subfigure}    
    \caption{Images with the highest cosine similarity of features between before and after our DAFT for each class.
    }
    \label{fig:cos_high-test-images-daft}
\end{figure}

\begin{figure}[htbp]
    \centering
        \begin{subfigure}[b]{0.99\columnwidth}         
         \includegraphics[width=\columnwidth]{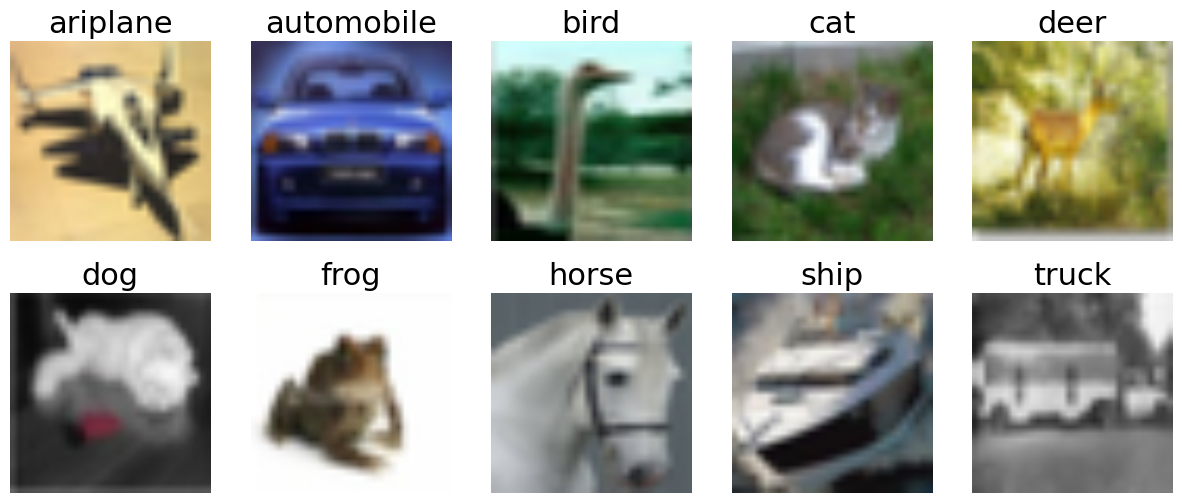}
         \caption{CIFAR-10 test images}
        \label{fig:cifar10-cos_low-test-images-daft}
        \end{subfigure}
    \medskip
        \begin{subfigure}[b]{0.99\columnwidth}         
         \includegraphics[width=\columnwidth]{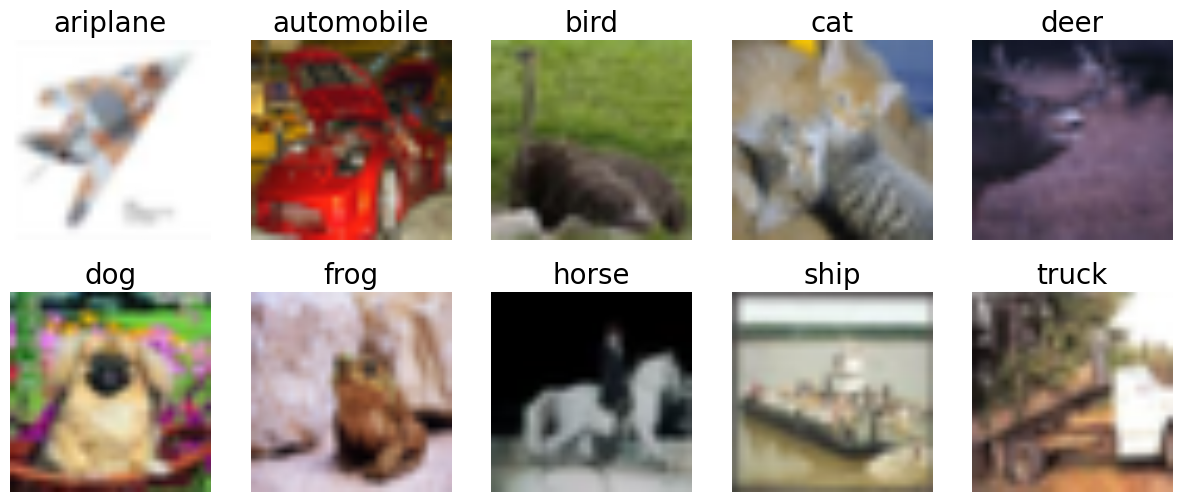}
         \caption{CIFAR-10.1 test images}
        \label{fig:ciafr10.1-cos_low-test-images-daft}
        \end{subfigure}
    \medskip     
        \begin{subfigure}[b]{0.99\columnwidth}         
         \includegraphics[width=\columnwidth]{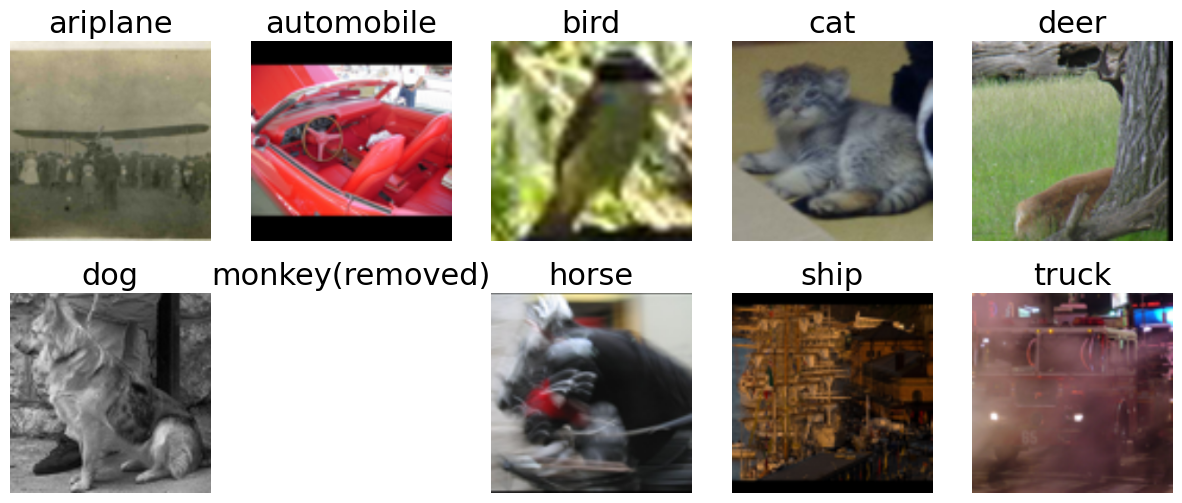}
         \caption{STL test images}
        \label{fig:STL-cos_low-test-images-daft}
        \end{subfigure}    
    \caption{Images with the lowest cosine similarity of features between before and after our DAFT for each class.
    }
    \label{fig:cos_low-test-images-daft}
\end{figure}

\clearpage
We extend the same visualization to our DAFT approach. Figure \ref{fig:cos_high-test-images-daft} and Figure \ref{fig:cos_low-test-images-daft} represent our results. Notably, images in Figure \ref{fig:cos_low-test-images-daft} exhibit fewer spurious features compared to those in Figure \ref{fig:cos_low-test-images}. This observation implies that our DAFT is less susceptible to the influence of spurious features on the feature extractor compared to FT.

\end{document}